\documentclass[journal]{IEEEtran}

\ifCLASSINFOpdf
\else
\fi

\usepackage{times}
\usepackage{epsfig}
\usepackage{graphicx}
\usepackage{amsmath}
\usepackage{amssymb}
\usepackage{graphicx}
\usepackage{cite}
\usepackage{enumerate}
\usepackage{cases}
\usepackage{multirow}
\usepackage{verbatim}
\usepackage{amssymb}
\usepackage{CJK}
\usepackage{algorithm}
\usepackage{algorithmicx}
\usepackage{algpseudocode}

\usepackage{color}
\usepackage{bm}
\usepackage{booktabs}
\usepackage{hyperref}

\begin{document}

\title{HSCS: Hierarchical Sparsity Based Co-saliency Detection for RGBD Images}


\author{Runmin Cong, Jianjun Lei,~\IEEEmembership{Senior Member,~IEEE,} Huazhu Fu,~\IEEEmembership{Senior Member,~IEEE,}\\ Qingming Huang,~\IEEEmembership{Fellow,~IEEE,} Xiaochun Cao,~\IEEEmembership{Senior Member,~IEEE,} and Nam Ling,~\IEEEmembership{Fellow,~IEEE}
\thanks{Manuscript received Feb. 2018. This work was supported in part by the National Natural Science Foundation of China under Grant 61520106002, Grant 61722112, Grant 61731003, Grant 61332016, Grant 61620106009, Grant U1636214, Grant 61602345, in part by the Key Research Program of Frontier Sciences, CAS under Grant QYZDJ-SSW-SYS013, and in part by the Technology Research and Development Program of Tianjin under Grant 15ZXHLGX00130. (\emph{Corresponding author: Jianjun Lei})}
\thanks{R. Cong and J. Lei are with the School of Electrical and Information Engineering, Tianjin University, Tianjin 300072, China (e-mail: rmcong@tju.edu.cn; jjlei@tju.edu.cn).}
\thanks{H. Fu is with the Inception Institute of Artificial Intelligence, Abu Dhabi, United Arab Emirates (e-mail: huazhufu@gmail.com).}
\thanks{Q. Huang is with the School of Computer and Control Engineering, University of Chinese Academy of Sciences, Beijing 100190, China (e-mail: qmhuang@ucas.ac.cn).}
\thanks{X. Cao is with State Key Laboratory of Information Security, Institute of Information Engineering, Chinese Academy of Sciences, Beijing 100093, China, and also with University of Chinese Academy of Sciences, Beijing 100190, China (e-mail: caoxiaochun@iie.ac.cn).}
\thanks{N. Ling is with the Department of Computer Engineering, Santa Clara University, Santa Clara, CA 95053, USA (e-mail: nling@scu.edu).}
}

\markboth{IEEE TRANSACTIONS ON MULTIMEDIA}%
{Shell \MakeLowercase{\textit{et al.}}: Bare Demo of IEEEtran.cls for IEEE Journals}

\maketitle

\begin{abstract}
Co-saliency detection aims to discover common and salient objects in an image group containing more than two relevant images. Moreover, depth information has been demonstrated to be effective for many computer vision tasks. In this paper, we propose a novel co-saliency detection method for RGBD images based on hierarchical sparsity reconstruction and energy function refinement. With the assistance of the intra saliency map, the inter-image correspondence is formulated as a hierarchical sparsity reconstruction framework. The global sparsity reconstruction model with a ranking scheme focuses on capturing the global characteristics among the whole image group through a common foreground dictionary. The pairwise sparsity reconstruction model aims to explore the corresponding relationship between pairwise images through a set of pairwise dictionaries. In order to improve the intra-image smoothness and inter-image consistency, an energy function refinement model is proposed, which includes the unary data term, spatial smooth term, and holistic consistency term. Experiments on two RGBD co-saliency detection benchmarks demonstrate that the proposed method outperforms the state-of-the-art algorithms both qualitatively and quantitatively.
\end{abstract}


\IEEEpeerreviewmaketitle

\section{Introduction}

\IEEEPARstart{V}{ISUAL} attention mechanism enables people to quickly locate the most interesting parts or salient objects from a complex scene. As a branch of computer vision, saliency detection is devoted to enabling a computer to discover these salient regions automatically. This process has been applied in a large number of visual tasks, such as segmentation \cite{R1,R6,R7}, retargeting \cite{R3}, enhancement \cite{R4}, foreground annotation \cite{R5}, and blur detection \cite{R2}. The last decade has witnessed the vigorous development and qualitative leap in performance of image saliency detection \cite{R8,R10,R63,R9,R11,R12,R13,R15,R81,R19,R20,R22,R23,R71,R72,R76,R79,R82,R83}.\par

With the recent explosive growth of data volume, people need to process multiple relevant images collaboratively. As an extension of traditional image saliency detection, co-saliency detection aims to discover common and salient objects in an image group containing multiple relevant images \cite{R24}. This has been successfully applied in co-segmentation \cite{R25,R26,R68}, co-localization \cite{R27}, and image matching \cite{R29}. Different from image saliency detection, the co-saliency detection model needs to consider the common attributes of salient objects in an image group through the inter-image constraint. In other words, the co-salient objects should not only be prominent with respect to the backgrounds in each individual image, but also should recur throughout the whole image group.\par

In addition to color appearance, humans can perceive the distance mapping of a scene, which is known as depth information. With the development of imaging devices and technologies, capturing the depth representation information becomes increasingly convenient. Moreover, depth information has been proven to be useful for many computer vision tasks, such as segmentation \cite{R32}, enhancement \cite{R31}, and saliency detection \cite{R33,R34,R35,R37,R38,R39,R69,R80}. However, most of the existing methods focus on handling the RGBD images rather than the RGBD image group. In this paper, the depth feature is not only served as a constraint in the inter-image correspondence modelling, but also used as a color information supplement in the refinement component.\par

The corresponding relationship among multiple images plays an important role in co-saliency detection. In other words, in addition to the saliency attribute in an individual image, the repetitiveness constraint across the whole image group is also crucial to suppress the background and non-common salient regions. In existing methods, the inter-image correspondence is simulated as a matching process \cite{R40,R41,R42,R43,R44,R75}, clustering process \cite{R45,R66,R78}, low-rank problem \cite{R46,R47}, propagation process \cite{R48,R49,R65}, or learning process \cite{R50,R51,R52,R70,R53}. However, the matching- and propagation-based methods are often time consuming, while the clustering based methods are sensitive to the noise. To overcome these problems, the sparsity-based technique is a good choice and has demonstrated the potential to improve the performance of many tasks, including saliency detection \cite{R30,R54,R55,R56}. For the sparsity-based saliency detection methods, the background or foreground dictionary is used to reconstruct each processing unit, and the saliency is measured by the reconstruction error. In addition to describing the saliency of an individual image, sparsity representation can be used to constrain the inter-image correspondence capturing and to achieve inter saliency detection. In this paper, a hierarchical sparsity reconstruction model is innovatively proposed to capture a more comprehensive inter-image relationship by considering the global and local inter-image information. The hierarchical sparsity property includes two complementary aspects, i.e., (1) The co-salient objects in the whole image group should belong to the same category and have similar appearance. Therefore, a global foreground dictionary with a ranking scheme is built to reconstruct each image and to capture the global inter-image correspondence, which is called global sparsity reconstruction. (2) The relationship among multiple images can be decomposed into a combination of multiple pairwise correspondences. Therefore, a set of foreground dictionaries constructed by other images are utilized to reconstruct the current image and obtain multiple pairwise inter saliency maps from the local perspective.\par

The co-salient objects in different images of the same group should be similar and consistent in appearance. Thus, a superior co-saliency detection model should guarantee the local smoothness in each individual image and global consistency in the whole image group. In this paper, we propose an energy function refinement model to attain a more consistent and accurate co-saliency result, which includes the unary data term, spatial smooth term, and holistic consistency term. The data term constrains the updating degree of the refinement algorithm, and the smooth term favors that all the spatially adjacent regions with similar appearance should be assigned to consistent saliency scores. In addition to these two traditional terms, a holistic consistency term is specifically designed for the co-saliency detection task, which imposes the appearances of co-salient objects to be consistent in the whole image group.\par

In this paper, we provide an effective and efficient co-saliency detection method for RGBD images based on hierarchical sparsity reconstruction and energy function refinement. The main contributions are summarized as follows:\par

\begin{enumerate}[(1)]
\item A co-saliency detection method for RGBD images is proposed that integrates the intra saliency detection, hierarchical inter saliency detection based on global and pairwise sparsity reconstructions, and energy function refinement. The hierarchical sparsity representation is firstly used to capture the inter-image correspondence in co-saliency detection.
\item The global sparsity reconstruction is utilized to capture the global characteristic among the whole image group through a common foreground dictionary. Moreover, a ranking scheme is designed to guide the foreground seed selection and optimize the common foreground dictionary.
\item The inter-image relationship is simulated as a combination of multiple pairwise correspondences. The pairwise sparsity reconstruction model utilizes a set of foreground dictionaries produced by other images to explore local inter-image information.
\item To improve the intra-image smoothness and inter-image consistency, an energy function refinement model is proposed. A holistic consistency term is specifically designed for the co-saliency detection task, which constrains the appearances of co-salient objects to be consistent in the whole image group.
\end{enumerate}\par
The rest of the paper is organized as follows. Section II reviews related works. Section III presents the details of the proposed RGBD co-saliency detection method. The experimental comparisons and discussions are presented in Section IV. Finally, the conclusion is drawn in Section V.\par

\begin{figure*}[!t]
\centering
\includegraphics[width=1\linewidth]{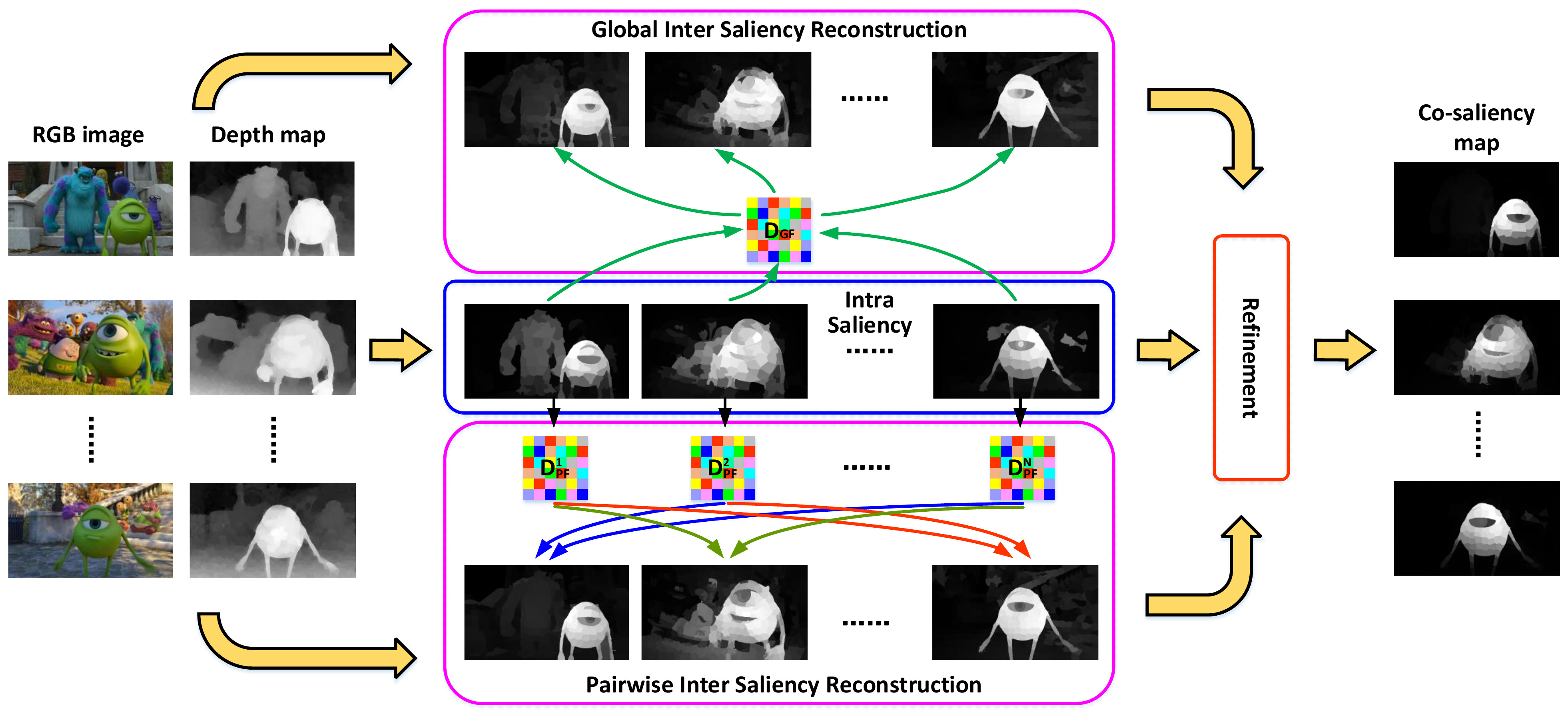}
\caption{The flowchart of the proposed RGBD co-saliency detection method.}
\label{fig1}
\end{figure*}

\section{Related Work}
In this section, we briefly review the related works of image saliency detection and co-saliency detection.\par
\subsection{Image Saliency Detection}
The last decade has witnessed the considerable development of saliency detection for RGB image, and a large number of methods have been presented \cite{R8,R10,R63,R9,R11,R12,R13,R15,R81,R19,R20,R22,R23,R71,R72,R76,R79,R82,R83}. Li \emph{et al.} \cite{R10} used the reconstruction error to measure the saliency of a region, where the salient region corresponds to a larger reconstruction error. Zhu \emph{et al.} \cite{R9} proposed a principled optimization framework integrating multiple low-level cues to achieve saliency detection with the help of a robust background measure. Peng \emph{et al.} \cite{R15} proposed a structured matrix decomposition based saliency detection method guided by high-level priors and obtained competitive performance. Recently, deep learning has been demonstrated the power in saliency detection. Li and Yu \cite{R19} proposed an end-to-end deep contrast network for saliency detection, which integrates the multi-scale fully convolutional stream and the segment-wise spatial pooling stream. Hou \emph{et al.} \cite{R22} introduced short connections into the skip-layer structures within the holisitcally-nested edge detector architecture to achieve image saliency detection. Zhang \emph{et al.} \cite{R23} utilized an encoder Fully Convolutional Network (FCN) and a corresponding decoder FCN to detect the salient object, in which the reformulated dropout and hybrid up-sampling are designed. In addition, some new measurements for saliency evaluation are proposed, such as Structure-measure \cite{R82} and Enhanced-alignment measure \cite{R83}.\par

In addition, the introduction of depth information makes RGBD images more in accordance with the human vision system, which could improve the performance of saliency detection \cite{R33,R34,R35,R37,R38,R39,R69,R80}. To achieve RGBD saliency detection, some depth features and measures are utilized. Ju \emph{et al.} \cite{R34} proposed an Anisotropic Center-Surround Difference (ACSD) measure to calculate the depth-aware saliency map. Considering the quality of depth map, Cong \emph{et al.} \cite{R37} proposed an RGBD saliency detection method via depth confidence analysis and multiple cues fusion, where the depth confidence measure works as a controller to constrain the introduction of depth information in the saliency model. In \cite{R38}, depth information is used as a regional feature for low-level contrast-based saliency computation, and also worked as a weighting term for mid-level saliency evaluation. Qu \emph{et al.} \cite{R39} designed a convolutional neural network to automatically learn the interaction between low-level cues and saliency result for RGBD saliency detection.\par

\subsection{Co-saliency Detection}
In order to achieve co-saliency detection, the inter-image correspondence can be captured by different techniques, such as similarity matching \cite{R40,R41,R42,R43,R44,R75}, clustering \cite{R45,R66,R78}, low-rank decomposition \cite{R46,R47}, propagation \cite{R48,R49,R65}, and learning \cite{R50,R51,R52,R70,R53}. Liu \emph{et al.} \cite{R43} proposed a novel co-saliency detection model integrating the global similarity on the fine segmentation level with the object prior on the coarse segmentation level, where the inter-image correspondence is formulated as global similarity of each region. Li \emph{et al.} \cite{R44} proposed a two-stage saliency model guided co-saliency detection method, in which the first stage recovers   the co-salient parts through the efficient manifold ranking, and the second stage captures the inter-image correspondence via a ranking scheme. In \cite{R45}, an efficient cluster-based co-saliency detection algorithm for multiple images is proposed, which takes the cluster as the basic unit to represent the multi-image relationship by integrating the contrast, spatial, and corresponding cues. Cao \emph{et al.} \cite{R46} proposed a saliency fusion framework for co-saliency detection based on the rank constraint, which is valid for multiple images and also works well on single image saliency detection. In \cite{R48}, co-saliency detection is formulated as a two-stage saliency propagation problem, including the intra-saliency propagation stage and the inter-saliency propagation stage. With the training process, the learning based methods always achieve competitive performance. Zhang \emph{et al.} \cite{R50} proposed a co-saliency detection model under the Bayesian framework with some higher-level features extracted by the convolutional neural network. Wei \emph{et al.} \cite{R51} proposed an end-to-end group-wise deep co-saliency detection model, where the semantic block is utilized to obtain the basic feature representation, the group-wise and single feature representation blocks are used to capture the group-wise interaction information and individual image information, and the collaborative learning structure with convolution-deconvolution model aims to output the co-saliency map. Han \emph{et al.} \cite{R53} introduced the metric learning into co-saliency detection to jointly learn the discriminative feature representation and co-salient object detector, which can handle the wide variation in image scene and achieve superior performance.\par

Furthermore, a few methods are proposed to achieve RGBD co-saliency detection by introducing the depth cue as an effective privileged information. Song \emph{et al.} \cite{R57} utilized the bagging-based clustering method to detect the co-salient objects from RGBD images group, where the average depth value, depth range, and Histogram of Oriented Gradient (HOG) on depth map are extracted to represent the depth attributes. Cong \emph{et al.} \cite{R58} proposed a co-saliency detection method for RGBD images by using the multi-constraint feature matching and cross label propagation, where the inter-image relationship is explored at the superpixel and image levels. In \cite{R59}, an iterative co-saliency detection framework for RGBD images is proposed, which integrates the addition scheme, deletion scheme, and iterative scheme. The addition scheme aims at generating the RGBD saliency map by introducing the depth shape prior into the existing saliency detection model, the deletion scheme focuses on capturing the inter-image correspondence via a common probability function, and the iterative scheme is served as an optimization process through a refinement-cycle.\par

Compared to the traditional saliency detection, there are three challenges for RGBD co-saliency detection, i.e., (1) how to utilize the depth information to enhance the identification of salient objects, (2) how to capture the corresponding relationship among multiple images, and (3) how to guarantee the consistency and smoothness of the final co-saliency map. To address these challenges, we propose a hierarchical sparsity based co-saliency detection method for RGBD images. The depth feature is used as an additional constraint and supplement cue in the inter-image correspondence and refinement components. In addition, an effective inter saliency model is designed to capture the inter-image correspondence, which integrates the global sparsity reconstruction and pairwise sparsity reconstruction. Moreover, we also formulate an energy function refinement to achieve a superior and globally consistent co-saliency map.\par

\section{Proposed Method}
\subsection{Framework}
The flowchart of the proposed hierarchical sparsity based co-saliency detection method for RGBD images is shown in Fig. \ref{fig1}, which includes intra saliency calculation, hierarchical inter saliency detection based on global and pairwise sparsity reconstructions, and energy function refinement.\par

According to the definition of co-saliency detection, the co-salient objects should be prominent in an individual image. Therefore, the intra saliency map is firstly calculated for each individual image. We denote the input RGB images in a group as $\{I^{i}\}_{i=1}^{N}$, and the corresponding depth maps as $\{D^{i}\}_{i=1}^{N}$, where $N$ is the number of images in the group. For computational efficiency and structural representation, each RGB image $I^{i}$ is abstracted into some superpixels $\textbf{R}^{i}=\{r^{i}_{m}\}_{m=1}^{N^{i}}$ through the SLIC algorithm \cite{R60}, where $N^{i}$ represents the number of superpixels in image $I^{i}$. In light of the effectiveness and robustness of the DCMC method \cite{R37}\footnote{ \url{https://rmcong.github.io/proj_RGBD_sal.html} }, we chose it as the basic method for intra saliency detection, and the intra saliency value of superpixel $r^{i}_{m}$ is denoted as $S_a(r^{i}_{m})$.\par

The background of each image may be diverse within the same image group, while the co-salient objects tend to have a similar appearance in all images. Therefore, the co-salient regions can be reconstructed better than the background regions through a sparsity framework with the foreground dictionary. In this paper, the corresponding relationship among multiple images is simulated as a hierarchical sparsity framework considering the global and pairwise sparsity reconstructions. The global inter saliency reconstruction model describes the inter-image correspondence from the perspective of the whole image group via a common reconstruction dictionary, while the pairwise inter saliency reconstruction model utilizes a set of foreground dictionaries produced by other images to capture local inter-image information.\par

Finally, an energy function refinement model, including the unary data term, spatial smooth term, and holistic consistency term, is proposed to improve the intra-image smoothness and inter-image consistency and to generate the final co-saliency map. The spatial smooth term is used to optimize the intra-image smoothness, and the holistic consistency term is specifically designed for co-saliency detection task to update the inter-image consistency. Hierarchical inter saliency detection based on global and pairwise sparsity reconstructions, as well as the energy function refinement, are detailed in the following sections.\par

\subsection{Global Inter Saliency Reconstruction}
The co-salient objects in a whole image group should belong to the same category and have a similar appearance. Therefore, a global foreground dictionary is built to reconstruct each image and capture the global inter-image correspondence. First, some initial foreground seeds are selected based on all the intra saliency maps in the image group. Then, a ranking filter is designed to eliminate the interference seeds and to determine the optimal foreground seeds. Next, the feature vectors of foreground seeds are extracted to construct the global foreground dictionary. Finally, the reconstruction error produced by the sparsity framework is utilized to measure the global inter saliency.\par
\subsubsection{Initial foreground seeds selection}
The intra saliency map provides effective single image saliency description. We assume that most of the co-salient objects can be included in these saliency maps. Thus, the top $K$ superpixels in image $I^i$ with larger intra saliency values are selected as the foreground seeds. Then, all these seeds from different images are combined into an initial foreground seed set $\Phi_{init}=\Phi^{1}_{init}\cup\Phi^{2}_{init}\cup\ldots\Phi^{N}_{init}$, where $\Phi^{i}_{init}$ denotes the foreground seed set of image $I^i$, and $N$ is the number of images. \par

\subsubsection{Ranking based seeds filtering}
Since the intra saliency result is not completely accurate, some disturbances may be wrongly included in the initial foreground seed set, such as the backgrounds and non-common salient regions, which may degenerate the reconstruction accuracy. Therefore, we designed a ranking scheme to filter the interferences and refine the foreground seeds.\par

In general, the co-salient objects satisfy three constraints, i.e., (a) the category should be same, (b) the color appearance should be similar, and (c) the depth distribution should be approximate. Combining these three constraints, a novel measure is designed to evaluate the local consistency of superpixels belonging to the initial foreground seed set. First, all the initial seed superpixels are grouped into five clusters by using the K-means++ clustering \cite{R61}, and each superpixel is assigned to a corresponding cluster center $\{\emph{\textbf{c}}_i\}_{i=1}^{N\cdot K}$, respectively. Then, introducing the clustering, color, depth, and saliency constraints, the consistency measure is defined as follows:
\begin{equation}
mc(r_{m})=[\sum_{\substack{n=1\\ n\neq m}}^{N\cdot K} (1-\|\emph{\textbf{c}}_m-\emph{\textbf{c}}_n\|_2)\cdot \omega_{mn}]\cdot S_{a}(r_m)
\end{equation}
and
\begin{equation}
\omega_{mn}= \exp(-\frac{\chi^2(\emph{\textbf{h}}_m,\emph{\textbf{h}}_n)+\lambda_{min}\cdot |d_m-d_n|}{\sigma^2})
\end{equation}
where $r_m,r_n\in\Phi_{init}$; $\emph{\textbf{c}}_m$ is the cluster center of superpixel $r_m$; $\omega_{mn}$ represents the feature similarity between superpixel $r_m$ and superpixel $r_n$; $S_{a}(r_m)$ is the intra saliency value of superpixel $r_m$; $N\cdot K$ is the total number of initial foreground seeds; $\|\cdot\|_2$ is the $l_2$-norm function; and $\sigma^2$ is a parameter to control strength of the similarity, which is set to 0.1 in all experiments following \cite{R37}. $\emph{\textbf{h}}_m$ denotes the color histogram of superpixel $r_m$ in the Lab color space; $\chi^2(\cdot)$ represents the Chi-square distance; and $d_m$ is the mean depth value of superpixel $r_m$. $\lambda_{min}=\min(\lambda_m,\lambda_n)$ denotes the minimum depth confidence measure of two input depth maps, where $\lambda_{m}=\exp((1-m_m)\cdot CV_m\cdot H_m)-1$ is the depth confidence measure of the input depth map $D^m$; $m_m$ denotes the mean value of the whole depth image; $CV_m$ represents the coefficient of variation; and $H_m$ is the depth frequency entropy. More details can be found in \cite{R37}. A larger $mc$ corresponds to higher consistency with respect to other foreground seeds. In other words, the larger the consistency measure is, the higher the probability of the superpixel being the foreground seed. Finally, the top 80\% of initial seeds with larger consistency measure values are reserved as the final foreground seeds, which is denoted as $\Phi_{fin}$.\par

\begin{figure}[!t]
\centering
\includegraphics[width=1\linewidth]{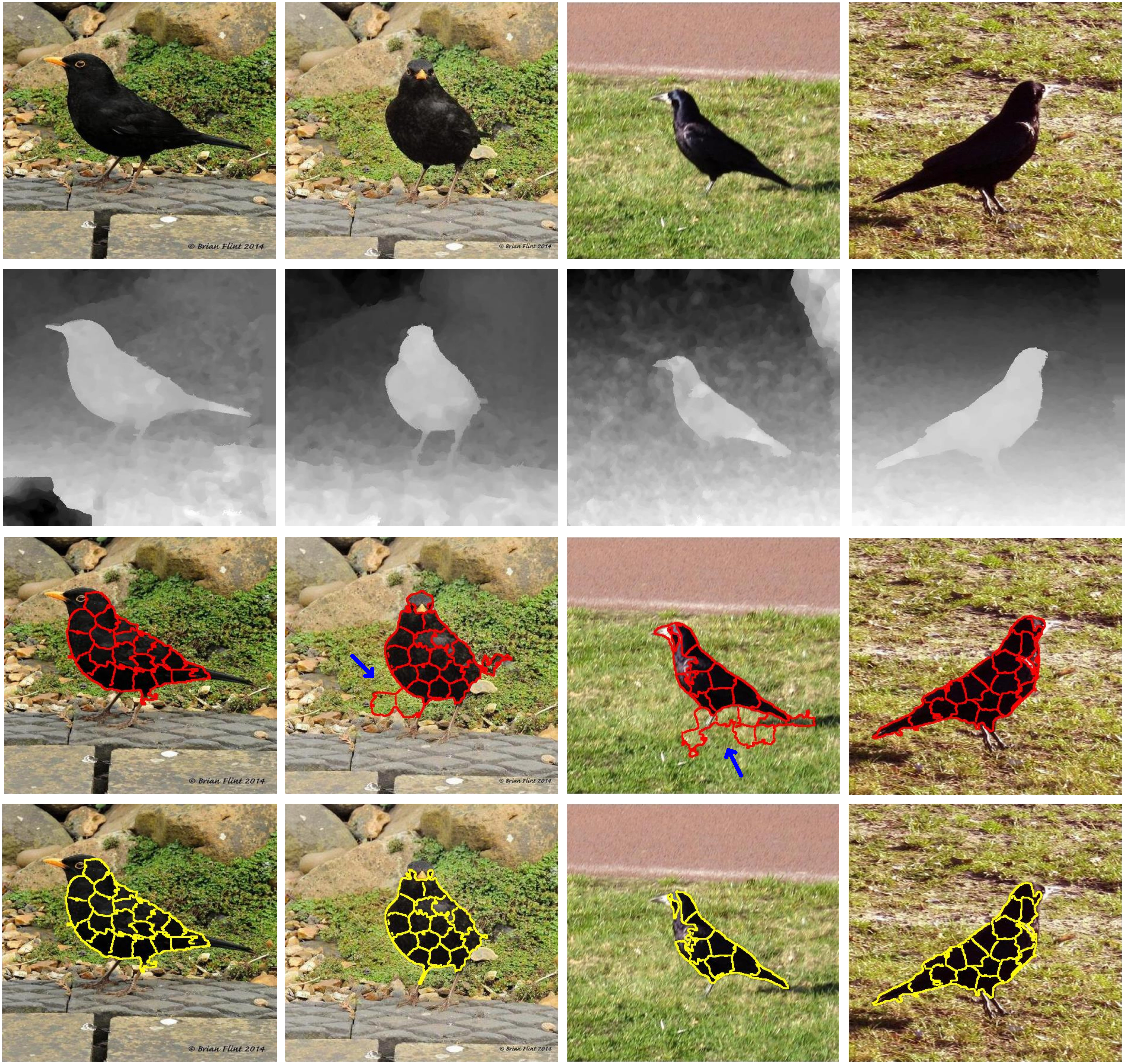}
\caption{Some examples of ranking scheme for foreground seeds selection. The first second rows are the RGB images and depth maps, the third row shows the initial foreground seeds marked in red, and last row presents the final foreground seeds marked in yellow after ranking scheme.
\label{fig2}}
\end{figure}

Some illustrations of the foreground seeds are shown in Fig. \ref{fig2}, where the third row presents the visualization of the initial foreground seeds marked in red, and the final foreground seeds marked in yellow after ranking scheme are shown in the last row. As can be seen, some backgrounds (e.g., the lawns located by the blue arrows) in the second and third images are wrongly selected as the foregrounds in the initial seeds set. With the ranking scheme, the correct foreground seeds (i.e., the dark bird) are successfully reserved, while the backgrounds are effectively eliminated.\par
\subsubsection{Sparsity-based global reconstruction}
Four types of low-level features, including color components, depth attribute, spatial location, and texture distribution, are utilized to describe each superpixel $r_m^i$ as $\emph{\textbf{f}}^{i}_{m}=[\emph{\textbf{l}}^{i}_{m}\ d^{i}_{m}\ \emph{\textbf{p}}^{i}_{m}\ \emph{\textbf{t}}^{i}_{m}]^T$, where $\emph{\textbf{l}}$ is the 9-dimensional color components in the RGB, Lab, and HSV color spaces; $d$ denotes the depth value; $\emph{\textbf{p}}$ corresponds to the 2-dimensional spatial coordinates; and $\emph{\textbf{t}}$ represents the 15-dimensional texton histogram \cite{R62}. The feature representations of the stacking superpixels in the final foreground seeds set $\Phi_{fin}$ are constructed as the global foreground dictionary, which is denoted as $\textbf{D}_{GF}$.\par

Under the same reconstruction dictionary, the reconstruction error between foreground and background regions should be different. Thus, the image saliency can be measured by the reconstruction error \cite{R10}. We compute the reconstruction error by the sparsity representation with a global foreground dictionary, and each superpixel $r_m^i$ is encoded by:
\begin{equation}
\alpha_{m}^{i*}=\mathop{\arg\min}_{\alpha_{m}^{i}} \|\textbf{\emph{f}}_{m}^{i}-\textbf{D}_{GF}\cdot \alpha_{m}^{i}\|_{2}^{2}+\xi\cdot \|\alpha_{m}^{i}\|_1
\label{eqn3}
\end{equation}
where $\alpha_{m}^{i*}$ is the optimal sparse coefficient for superpixel $r_m^i$; $[\textbf{D}_{GF}]_{L\times |\Phi_{fin}|}$ denotes the global foreground dictionary; $|\Phi_{fin}|$ represents the number of final foreground seeds; $L=27$ is the feature dimension of each superpixel in our work; $\textbf{\emph{f}}_{m}^{i}$ is the feature representation of superpixel $r_m^i$; $\|\cdot\|_1$ is the $l_1$-norm function; and $\xi$ is set to 0.01 as suggested in \cite{R10}.\par

The foreground dictionary is used to achieve global reconstruction, thus, the superpixel with the smaller reconstruction error should be assigned to a greater saliency value and vice versa. The global inter saliency of superpixel $r_m^i$ is defined as:
\begin{equation}
S_{gr}(r_{m}^{i})=\exp (-\varepsilon_{m}^{i}/\sigma^2)=\exp (-\|\textbf{\emph{f}}_{m}^{i}-\textbf{D}_{GF}\cdot \alpha_{m}^{i*}\|_{2}^{2}/\sigma^2)
\end{equation}
where $S_{gr}(r_{m}^{i})$ is the inter saliency of superpixel $r_m^i$ through the global reconstruction; $\varepsilon_{m}^{i}$ denotes the reconstruction error of superpixel $r_m^i$; and $\sigma^2$ is a weighted constant.\par

\subsection{Pairwise Inter Saliency Reconstruction}
The global reconstruction aims to describe the inter-image correspondence from the perspective of the whole image group. In fact, the relationship among multiple images can be decomposed into a combination of multiple pairwise correspondences, which benefits capturing the local inter-image information. In order to deeply explore a more comprehensive inter-image corresponding relationship, a sparsity-based pairwise reconstruction method is proposed to calculate the pairwise inter saliency. First, we construct a foreground dictionary for each image based on the corresponding intra saliency map, respectively. In this way, the $N$ foreground dictionaries in an image group are obtained, where $N$ denotes the number of images in the group. Then, each image is reconstructed by the $N-1$ foreground dictionaries derived from other images in the group, respectively. Finally, these $N-1$ reconstructed results are fused to generate the pairwise inter saliency map.\par

For each image $I^k$, the top $K$ superpixels with larger intra saliency values are selected as the foreground seeds. Similar to the sparsity-based global reconstruction, a 27-dimensional feature vector is used to represent each superpixel. Then, the feature representations of the stacking foreground superpixels in each image are constructed as the pairwise foreground dictionary, which is denoted as $\textbf{D}^k_{PF}$. As mentioned earlier, the foreground pairwise dictionaries generated by other images can be utilized to reconstruct the current image and capture the local inter-image relationship. Using the pairwise foreground dictionary $\textbf{D}^k_{PF}$ produced by the image $I^k$, the image $I^i$ can be constructed and the saliency is measured as:
\begin{equation}
S_{pr}^{k}(r_{m}^{i})=\exp (-\varepsilon_{m}^{k,i}/\sigma^2)=\exp (-\|\textbf{\emph{f}}_{m}^{i}-\textbf{D}_{PF}^k\cdot \alpha_{m}^{k,i*}\|_{2}^{2}/\sigma^2)
\end{equation}
where $S_{ir}^{k}(r_{m}^{i})$ is the inter saliency through the pairwise reconstruction using the dictionary $[\textbf{D}^k_{PF}]_{L\times K}$; $\varepsilon_{m}^{k,i}$ denotes the reconstruction error of superpixel $r_{m}^{i}$; $\alpha_{m}^{k,i*}$ is the optimal sparse coefficient of superpixel $r_{m}^{i}$; $k\in[1,2,\ldots,N]$, $k\neq i$ represents the index of pairwise foreground dictionary; and $\sigma^2$ is a weighted parameter. Therefore, we obtain $N-1$ saliency maps for each image through different pairwise dictionaries. At last, all these maps are fused to generate the final pairwise inter saliency map by:
\begin{equation}
S_{pr}(r_{m}^i)=\frac{1}{N-1}\cdot\sum_{\substack{k=1\\ k\neq i}}^{N} S_{pr}^{k}(r_{m}^{i}).
\end{equation}\par

The global inter saliency map describes the global inter-image correspondence from the whole image group, while the pairwise inter saliency map captures the local relationship from the pairwise images. Finally, these two inter saliency maps are combined as the hierarchical sparsity based inter saliency:
\begin{equation}
S_{r}(r_{m}^i)=\frac{1}{2}\cdot (S_{gr}(r_{m}^i)+S_{pr}(r_{m}^i))
\end{equation}
where $S_{r}(r_{m}^i)$ denotes the hierarchical sparsity based inter saliency of superpixel $r_{m}^{i}$.\par
\begin{figure*}[!t]
\centering
\includegraphics[width=1\linewidth]{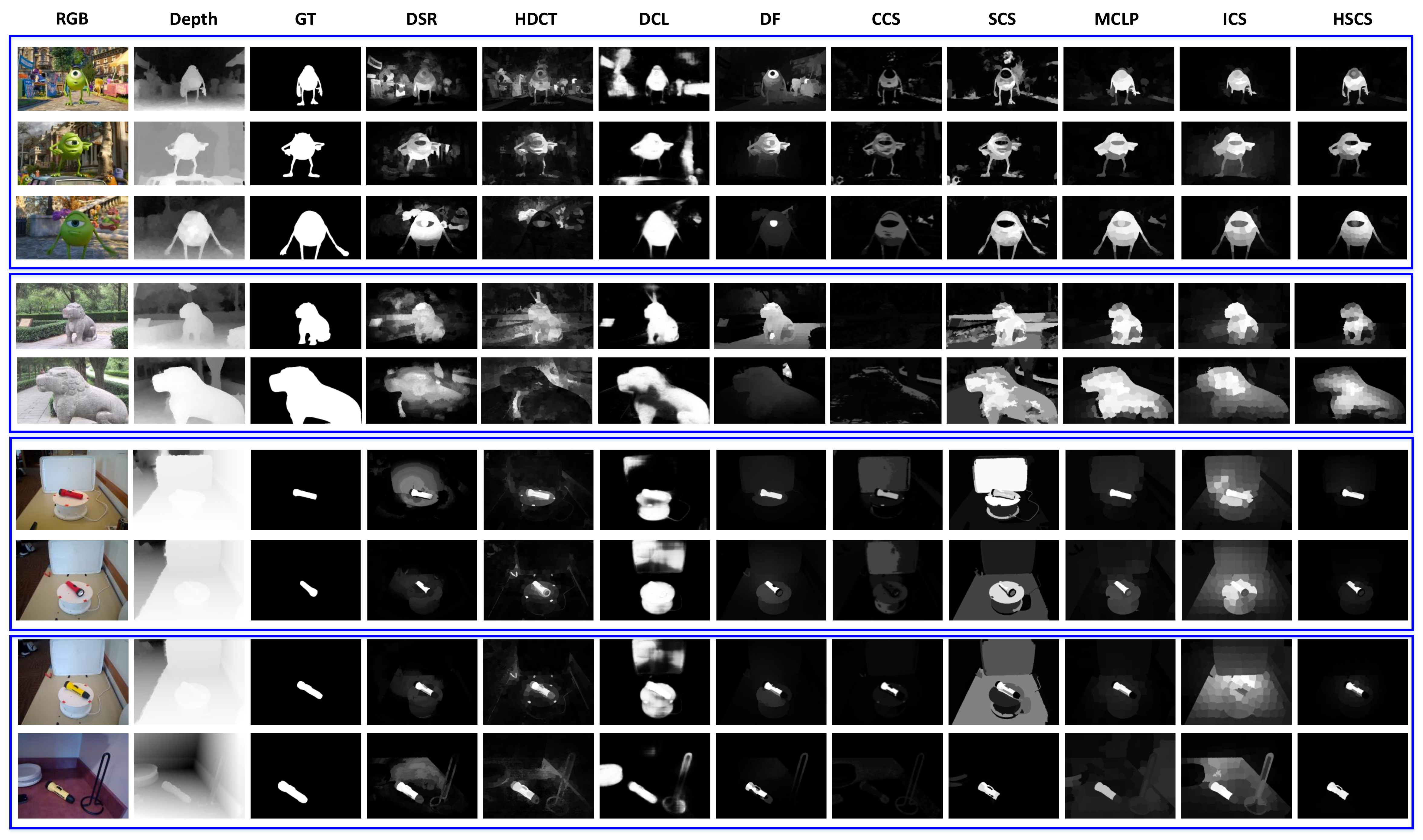}
\caption{Some visual examples of different methods.}
\label{fig3}
\end{figure*}

\subsection{Energy Function Refinement}
In order to achieve a superior and globally consistent saliency map, a refinement model with an energy function is designed in our work. Three terms are included in the energy function: the unary data term $T_u$ constrains the similarity between the final saliency map and initial saliency map; the spatial smooth term $T_s$ favors that all the similar and spatially adjacent superpixels in an individual image should be assigned to consistent saliency scores; and the holistic consistency term $T_h$ enforces that the appearance of the salient objects should be consistent within the whole image group. Therefore, the energy function is defined as:
\begin{equation}
\begin{aligned}
E&=T_u+T_s+T_h=\sum_{m} (\overline{s}_{m}-s_{m})^2\\
&+\sum_{(m,n)\in\Omega} \omega_{mn}\cdot (\overline{s}_{m}-\overline{s}_{n})^2+\sum_{m} g_{m}\cdot \overline{s}_{m}^2
\end{aligned}
\end{equation}
where $\overline{s}_{m}$ denotes the refined saliency value of superpixel $r_m$; $s_{m}=S_a(r_m)\cdot S_r(r_m)$ is the initial saliency value of superpixel $r_m$ by combining the intra and inter saliencies; $\Omega$ represents the spatially adjacent set in an individual image; $\omega_{mn}$ denotes the similarity between two superpixels, which is defined in the same way as Eq. (2); and $g_m=\chi^2(\emph{\textbf{h}}_m,\emph{\textbf{h}}_g)$ is the color difference between the superpixel $r_m$ and global foreground model via the chi-square distance of Lab color histograms. The top 20 superpixels with larger initial saliency value in each image are regarded as the foreground samples to represent the global foreground distribution.\par

Let $\textbf{\emph{s}}=[s_m]_{\aleph\times 1}$, and $\overline{\textbf{\emph{s}}}=[\overline{s}_m]_{\aleph\times 1}$, where $\aleph=\sum_{i=1}^{N} N_i$ is the total number of superpixels in the whole image group. Then, the energy function is rewritten in the matrix forms as:
\begin{equation}
\textbf{E}=(\overline{\textbf{\emph{s}}}-\textbf{\emph{s}})^T\cdot (\overline{\textbf{\emph{s}}}-\textbf{\emph{s}})+\overline{\textbf{\emph{s}}}^T\cdot (\textbf{D}-\textbf{W})\cdot \overline{\textbf{\emph{s}}}+\overline{\textbf{\emph{s}}}^T\cdot \textbf{G}\cdot \overline{\textbf{\emph{s}}} 
\end{equation}
where $\textbf{W}=[\omega_{mn}]_{\aleph\times \aleph}^{(m,n)\in\Omega}$ is the spatial color similarity matrix; $\textbf{D}=diag(d_1,d_2,\ldots,d_{\aleph})$ represents the degree matrix; $d_i=\sum_{j=1,(i,j)\in\Omega}^{\aleph} \omega_{ij}$; and $\textbf{G}=diag(g_1,g_2,\ldots,g_{\aleph})$ is the difference matrix between the superpixels and global foreground model.\par

The minimization of the above energy function can be solved by setting the derivative with respect to $\overline{\textbf{\emph{s}}}$ to be 0, which is represented as:
\begin{equation}
\frac{\partial \textbf{E}}{\partial \overline{\textbf{\emph{s}}}}= 2(\overline{\textbf{\emph{s}}}-\textbf{\emph{s}})+2(\textbf{D}-\textbf{W})\cdot \overline{\textbf{\emph{s}}}+2\textbf{G}\cdot \overline{\textbf{\emph{s}}}=0
\end{equation}\par
Combining the like terms, the solution is given by:
\begin{equation}
\overline{\textbf{\emph{s}}}=[\textbf{I}+(\textbf{D}-\textbf{W})+ \textbf{G}]^{-1}\cdot \textbf{\emph{s}}
\end{equation}
where $\textbf{I}$ is an identity matrix with the size of ${\aleph\times \aleph}$.\par

\section{Experiments}
In this section, we evaluate the proposed RGBD co-saliency detection method on the RGBD CoSal150 dataset and RGBD CoSeg183 dataset. The qualitative and quantitative comparisons with some state-of-the-art methods are presented, and some discussions and analyses are conducted.\par
\subsection{Experimental Settings}
In experiments, two public RGBD co-saliency detection datasets, named RGBD CoSal150 dataset\footnote{ \url{https://rmcong.github.io/proj_RGBD_cosal.html} } and RGBD CoSeg183 dataset\footnote{ \url{ http://hzfu.github.io/proj_rgbdseg.html} }, are used to evaluate the effectiveness of the proposed method. The RGBD CoSal150 dataset \cite{R58} consists of 150 RGBD images that are distributed in 21 indoor and outdoor scenes, and the pixel-level ground truth for each image is provided. The RGBD CoSeg183 dataset \cite{R25} is a challenging dataset due to the cluttered backgrounds, complex foreground patterns, and multiple objects, containing 183 RGBD images with pixel-level ground truth that are distributed in 16 indoor scenes. In our work, the number of superpixels for each image is set to 400, and the number of initial foreground seeds is set to $K=40$. The project is available on our website\footnote{ \url{https://rmcong.github.io/proj_RGBD_cosal_HSCS_tmm.html} }.\par

For quantitative evaluation, three criteria including the Precision-Recall (PR) curve, F-measure, and Mean Absolute Error (MAE) are introduced. The precision and recall scores are computed by comparing the binary saliency map against the ground truth, where the precision score represents the percentage of salient pixels which are allocated correctly in the obtained saliency map, and the recall value corresponds to the ratio of detected salient pixels with respect to the salient pixels in the ground truth. Thus, the PR curve can be drawn using the pairwise precision and recall scores. F-measure \cite{R84,R67} is an overall performance measurement, which is defined as the weighted mean of precision and recall:
\begin{equation}
F_{\beta}=\frac{(1+\beta^{2})Precision\times Recall}{\beta^{2}\times Precision+ Recall}
\end{equation}
where $\beta^{2}$ is set to 0.3 for emphasizing the precision as suggested in \cite{R64}.\par
Mean Absolute Error (MAE) is calculated as the average pixel-wise difference between the obtained saliency map $S$ and ground truth $G$ \cite{R74,R77}:
\begin{equation}
MAE=\frac{1}{W\times H} \sum_{i=1}^{W} \sum_{j=1}^{H} |S(i,j)-G(i,j)|
\end{equation}
where $W$ and $H$ are the width and height of the image, respectively.\par
\subsection{Comparison with State-of-the-art Methods}

\begin{figure*}[!t]
\centering
\includegraphics[width=0.9\linewidth]{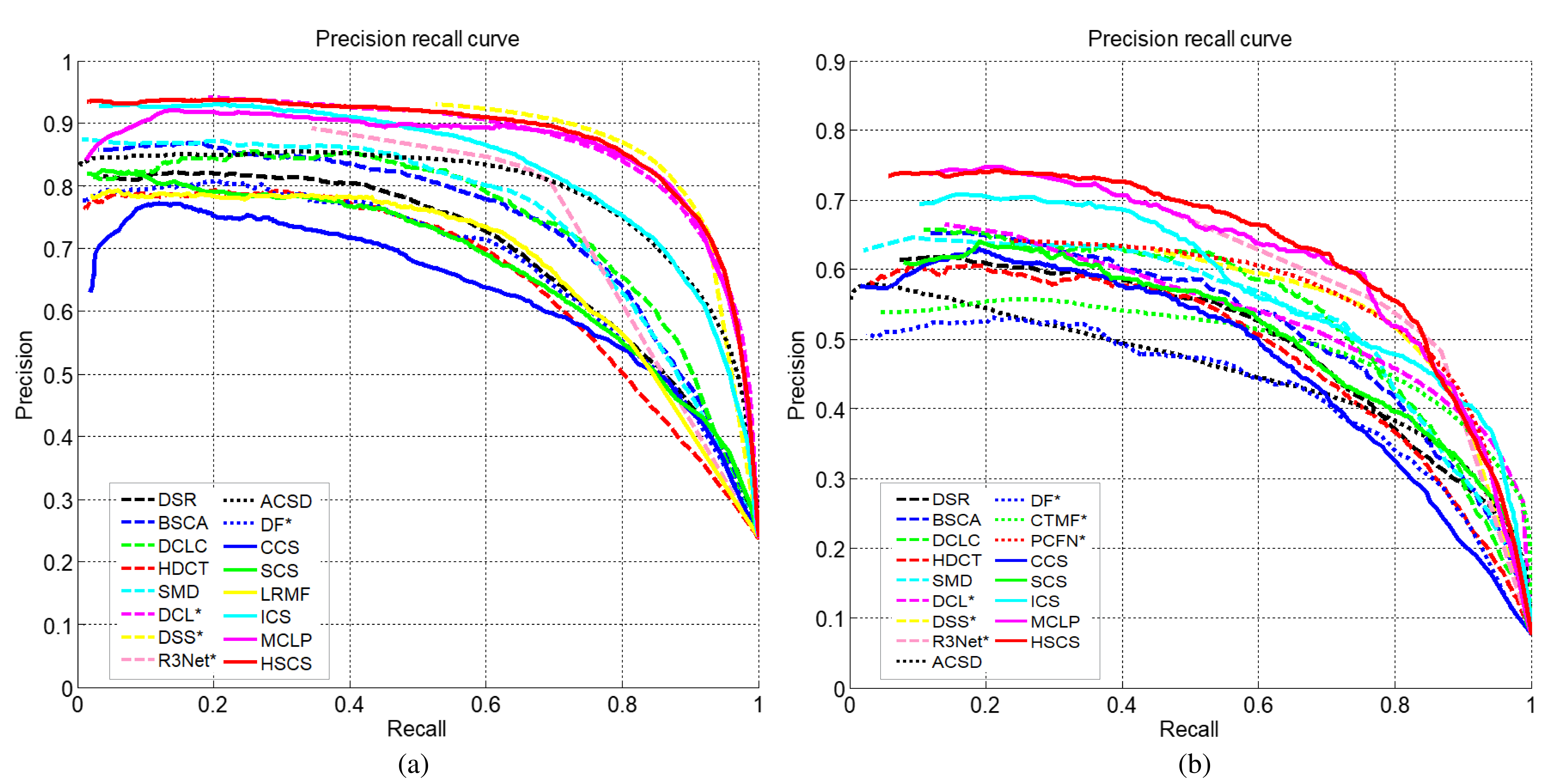}
\caption{PR curves of different methods on two RGBD co-saliency detection datasets, where ``*'' denotes the deep learning based methods. (a). RGBD CoSal150 dataset. (b). RGBD CoSeg183 dataset. }
\label{fig4}
\end{figure*}

We compared the proposed HSCS method with 17 state-of-the-art methods, including DSR \cite{R10}, BSCA \cite{R63}, DCLC \cite{R11}, HDCT \cite{R13}, SMD \cite{R15}, DCL \cite{R19}, DSS \cite{R22}, R3Net \cite{R72}, ACSD \cite{R34}, DF \cite{R39}, CTMF \cite{R69}, PCFN \cite{R80}, SCS \cite{R44}, CCS \cite{R45}, LRMF \cite{R47}, ICS \cite{R58}, and MCLP \cite{R59}, where DCL, DSS, R3Net, DF, CTMF, and PCFN are the deep learning based methods. The visual comparisons are shown in Fig. \ref{fig3}, and the quantitative evaluations are reported in Fig. \ref{fig4} and Table \ref{tab1}.\par

In Fig. \ref{fig3}, four image groups, including the green cartoon in the virtual scene, sculpture in the outdoor scene, and the red and yellow flashlights in the indoor scene, are illustrated for visual comparison. Due to the lack of a high-level feature description and inter-image constraints, the unsupervised single image saliency detection methods (e.g., DSR \cite{R10}, HDCT \cite{R13}) only roughly highlight the salient regions , while the background regions cannot be suppressed effectively (such as the street in the green cartoon group and the trees in the sculpture group). Benefitting from the strong learning ability of deep learning, the DCL \cite{R19} method achieves better performance with more consistent salient regions. However, there are still some wrongly detected backgrounds, such as the white object in the second image of the last group. Combining the depth cue and deep learning, the DF \cite{R39} method suppresses the background effectively, but it ignores the completeness of salient objects, such as the third image in the green cartoon group. For the RGB co-saliency detection methods (CCS \cite{R45} and SCS \cite{R44}), some foregrounds (such as the third image in the green cartoon group) are wrongly suppressed by the CCS method, and some backgrounds (such as the white board in the red flashlight group) are also inaccurately highlighted by the SCS method. Compared with the above methods, RGBD co-saliency detection methods (ICS \cite{R58} and MCLP \cite{R59}) achieve relatively superior performance with tangible salient objects. However, they still fail to suppress some common backgrounds, such as the ground in the sculpture group and the white board in the red flashlight group. By contrast, benefitting from the hierarchical reconstruction and global refinement, the proposed method can consistently highlight the salient objects and effectively suppress the backgrounds.\par

\begin{table}[!t]
\renewcommand\arraystretch{1.4}
\caption{Quantitative Comparisons with Different Methods on Two Datasets, Where ``*'' Denotes The Deep Learning Based Methods.}
\begin{center}
\setlength{\tabcolsep}{1.5mm}{
\begin{tabular}{c|c|c||c|c}
\toprule[2.2pt]
\multirow{2}{*}{} & \multicolumn{2}{c||}{RGBD CoSal150 Dataset} & \multicolumn{2}{c}{RGBD CoSeg183 Dataset}\\[0.5ex]
\cline{2-5}
  & F-measure & MAE & F-measure & MAE\\
\hline\hline
DSR \cite{R10} & $0.6956$ & $0.1867$ & $0.5496$ & $0.1092$ \\

BSCA \cite{R63} & $0.7318$ & $0.1925$ & $0.5678$ & $0.1877$ \\

DCLC \cite{R11} & $0.7385$ & $0.1728$ & $0.5994$ & $0.1097$ \\

HDCT \cite{R13} & $0.6753$ & $0.2146$ & $0.5447$ & $0.1307$ \\

SMD \cite{R15} & $0.7494$ & $0.1774$ & $0.5760$ & $0.1229$ \\

DCL* \cite{R19} & $0.8345$ & $0.1056$ & $0.5531$ & $0.0967$ \\

DSS* \cite{R22} & $0.8540$ & $0.0869$ & $0.5972$ & $0.0782$ \\

R3Net* \cite{R72} & $0.7812$ & $0.1296$ & $0.6190$ & $0.0678$ \\

ACSD \cite{R34} & $0.7788$ & $0.1806$ & $0.4787$ & $0.1940$ \\

DF* \cite{R39} & $0.6844$ & $0.1945$ & $0.4840$ & $0.1077$ \\

CTMF* \cite{R69} & $-$ & $-$ & $0.5316$ & $0.1259$ \\

PCFN* \cite{R80} & $-$ & $-$ & $0.6049$ & $0.0782$ \\

CCS \cite{R45} & $0.6311$ & $0.2138$ & $0.5383$ & $0.1210$ \\

SCS \cite{R44} & $0.6724$ & $0.1966$ &$0.5553$ & $0.1616$ \\

LRMF \cite{R47} & $0.6995$ & $0.1813$ & $-$ & $-$ \\

ICS \cite{R58} & $0.7915$ & $0.1790$ & $0.6011$ & $0.1544$ \\

MCLP \cite{R59} & $0.8403$ & $0.1370$ & $0.6365$ & $0.0979$ \\

HSCS & $0.8500$ & $0.1030$ & $0.6466$ & $0.0787$ \\
\bottomrule[2.2pt]
\end{tabular}}
\end{center}
\label{tab1}
\end{table}

PR curves of different methods on two datasets are shown in Fig. \ref{fig4}. As can be seen, the proposed HSCS method reaches a higher precision on the whole PR curves. Moreover, the proposed method is even superior to some deep learning based methods (e.g., DCL \cite{R19}, R3Net \cite{R72}, DF \cite{R39}, CTMF \cite{R69}, and PCFN \cite{R80}). The quantitative measurements, including F-measure and MAE score, are reported in Table \ref{tab1}. From the table, we can see that the proposed method achieves the competitive performance compared with 17 other state-of-the-art methods. On the RGBD CoSal150 dataset, the F-measure of the proposed method reaches 0.8500, and the maximum percentage gain reaches 34.7\% compared with other methods. Especially, the proposed HSCS method also achieves the percentage gain of 8.8\% compared with the deep learning based method (e.g., R3Net \cite{R72}). On the RGBD CoSeg183 dataset, the proposed method achieves the best performance in terms of F-measure, and the performance gains against others are more remarkable. The maximum percentage gain of the proposed method also reaches 35.1\% in terms of F-measure. All these visual examples and quantitative measures demonstrate the effectiveness of the proposed method.\par

\subsection{Discussions}
In this section, we conduct some discussions, including the module analysis, depth and ranking scheme evaluation, parameter discussion, and running time.\par
\subsubsection{\textbf{Module Analysis}}
The key points of the proposed hierarchical sparsity-based co-saliency detection method for RGBD images include a hierarchical sparsity based inter saliency model and an energy function refinement model. For hierarchical sparsity based inter saliency generation, the global and pairwise sparsity reconstructions are used to capture the inter-image constraints from two aspects. We comprehensively evaluate each module on the RGBD CoSal150 dataset, and the F-measures are presented in Table \ref{tab2}. The global inter saliency reconstruction captures the global corresponding relationship throughout the whole image group and achieves the F-measure of 0.8145. As a supplement, the multiple images relationship is formulated as pairwise correspondences by using the pairwise reconstruction model with a set of pairwise dictionaries, and the F-measure reaches 0.7628. Combining these two aspects, the hierarchical inter saliency structure can explore a more comprehensive inter-image relationship, and reaches 0.8198 in terms of F-measure, which is superior to most of the existing (co-)saliency detection methods (e.g., DSR \cite{R10}, SMD \cite{R15}, DF \cite{R39}, SCS \cite{R44}, LRMF \cite{R47}, and ICS \cite{R58}). Finally, the co-saliency detection with energy function refinement achieves the best performance, and the percentage gain reaches 3.7\% compared with the inter saliency models. \par 
\begin{table}[!t]
\renewcommand\arraystretch{1.5}
\caption{F-measure of The Main Modules on The RGBD CoSal150 Dataset.}
\begin{center}
\setlength{\tabcolsep}{5mm}{
\begin{tabular}{c|c}
\toprule[1.2pt]
Modules & F-measure \\[0.5ex]
\hline\hline
intra saliency detection & $0.8348$ \\
global reconstruction & $0.8145$ \\
pairwise reconstruction & $0.7628$ \\
hierarchical inter saliency & $0.8198$ \\
energy function refinement & $0.8500$ \\
\bottomrule[1.2pt]
\end{tabular}}
\end{center}
\label{tab2}
\end{table}

\begin{table}[!t]
\renewcommand\arraystretch{1.5}
\caption{Evaluation of Depth and Ranking Scheme on the RGBD CoSal150 Dataset, Where ``w/o'' Means ``without'' and ``w/'' Corresponds to ``with''.}
\begin{center}
\setlength{\tabcolsep}{6mm}{
\begin{tabular}{c|c}
\toprule[1.2pt]
 & F-measure \\[0.5ex]
\hline\hline
w/o depth and w/ ranking & $0.7839$ \\
w/ depth and w/o ranking & $0.8439$ \\
w/ depth and w/ ranking & $0.8500$ \\
\bottomrule[1.2pt]
\end{tabular}}
\end{center}
\label{tab3}
\end{table}
\subsubsection{\textbf{Depth and ranking scheme evaluation}}
In this paper, the depth cue is not only served as a constraint in the inter-image correspondence modelling, but also used as a supplement of color information in the refinement component. In order to attain more robust and accurate foreground seeds for global dictionary construction, a ranking scheme is designed to filter the interferences and to obtain optimal foreground seeds. We conduct some experiments on the RGBD CoSal150 dataset to evaluate the influence of these two constraints, and the F-measures are reported in Table \ref{tab3}. Compared with the first and the third rows, introducing the depth cue into the model, the performance is obviously improved with a percentage gain of 8.4\%. Shown in the second and third rows, the performance with the ranking scheme is better than the model without the ranking scheme. In addition, some illustrations are shown in Fig. \ref{fig2}. As can be seen, with the ranking scheme, the correct foreground seeds are successfully reserved, while the backgrounds (such as the lawn in the second and third images) are effectively eliminated. All these data demonstrate the effectiveness of the depth information and ranking scheme.\par

\begin{figure}[!t]
\centering
\includegraphics[width=0.8\linewidth]{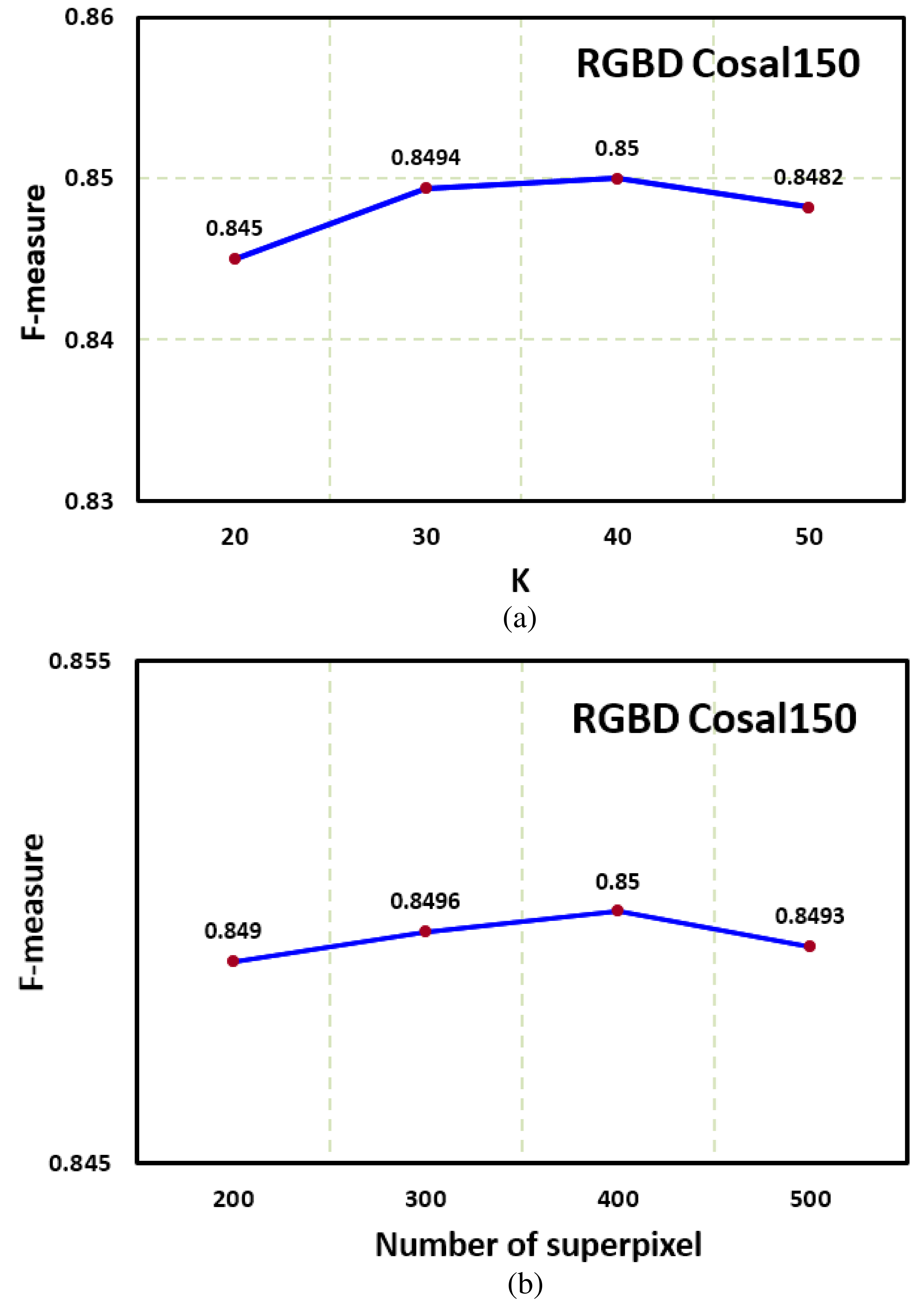}
\caption{The F-measure of different parameters on RGBD Cosal150 dataset. (a) The influence of different number of initial foreground seeds. (b) The influence of different number of superpixels.
\label{fig5}}
\end{figure}

\begin{table}[!t]
\renewcommand\arraystretch{1.5}
\caption{Comparisons of the average running time (seconds per image) on the RGBD CoSal150 Dataset.}
\begin{center}
\setlength{\tabcolsep}{1.5mm}{
\begin{tabular}{c|c|c|c|c|c|c|c|c}
\toprule[1.2pt]
Method & DCLC & SMD & DF & CCS & SCS & MCLP & ICS & HSCS \\[0.5ex]
\hline
Time & $1.96$ & $7.49$ & $12.95$ & $2.65$ & $2.94$ & $41.03$ & $42.67$ & $8.29$ \\
\bottomrule[1.2pt]
\end{tabular}}
\end{center}
\label{tab4}
\end{table}

\subsubsection{\textbf{Parameter discussion}}
In this section, we mainly discuss the influence of different numbers of initial foreground seeds and superpixels. In the experiment, we evaluate one parameter by fixing the other parameters. The tendency chart of the F-measure is shown in Fig. \ref{fig5}. From Fig. \ref{fig5}(a), selecting 20 initial foreground seeds for each image is not enough to represent the common saliency attributes completely and degenerates the inter reconstruction result. As the seed number increase, the performance improves and reaches the optimum when $K$ is set to 40. When $K$ reaches 50, the performance of the algorithm begins to drop. The main reason for the drop after 50 is that too many seeds contain background regions and decrease the reconstruction accuracy. As mentioned above, the performance is not highly sensitive to the parameter $K$, and we set it to 40 in all experiments. In addition to the number of initial foreground seeds, we further discuss the influence of different numbers of superpixels in the experiments. From the curve shown in Fig. \ref{fig5}(b), when the number of superpixels is set to 400, the result achieves the best performance. In fact, the performance in different numbers of superpixels are similar, indicating that the proposed algorithm is insensitive to the number of superpixels.\par

\subsubsection{\textbf{Running time}}
We compare the running time of the proposed method with others on a Quad Core 3.7GHz workstation with 16GB RAM and implemented using MATLAB 2014a. The average running time is listed in Table \ref{tab4}. In general, compared with the image saliency detection method, co-saliency detection algorithm often requires more computation time, especially for the matching based methods (such as MCLP \cite{R59}, ICS \cite{R58}). For the three RGBD co-saliency detection methods, under the same conditions, the MCLP method takes 41.03 seconds for one image, the ICS method takes 42.67 seconds, and the proposed HSCS method takes an average of 8.29 seconds to process one image. Since the commonly used superpixel-level matching process is replaced by the hierarchical sparsity based reconstruction to capture the inter-image correspondence, the computational efficiency of the proposed algorithm is clearly improved.\par

\section{Conclusion}
In this paper, a novel co-saliency detection method for RGBD images based on hierarchical sparsity reconstruction and energy function refinement is proposed. The major contribution lies in the hierarchical sparsity based inter saliency modelling, where the global inter-image model with a ranking scheme is used to capture the global characteristic among the whole image group through a common foreground dictionary, and the pairwise inter-image model is devoted to exploring the local corresponding relationship through a set of pairwise foreground dictionaries. In addition, an energy function refinement model is proposed to further improve the intra-image smoothness and inter-image consistency. The comprehensive comparisons and discussions on two RGBD co-saliency detection datasets have demonstrated that the proposed method outperforms other state-of-the-art methods qualitatively and quantitatively.\par

\ifCLASSOPTIONcaptionsoff
  \newpage
\fi

{
\bibliographystyle{IEEEtran}
\bibliography{ref}
}

%








\end{document}